\documentclass[twocolumn]{article}

\usepackage[a4paper, textwidth=7.2in, top=1in, bottom=1in]{geometry}

\usepackage[utf8]{inputenc}
\usepackage[round]{natbib}
\usepackage[colorlinks=true, allcolors=blue, hypertexnames=false]{hyperref}
\usepackage[font=small, labelfont=bf]{caption}

\usepackage{graphicx}    
\usepackage{amsmath}     
\usepackage{amsfonts}    
\usepackage{booktabs}    
\usepackage{microtype}   
\usepackage[misc]{ifsym} 

\usepackage{authblk}

\usepackage{orcidlink}

\title{Delving into LLM-assisted writing in biomedical publications through excess vocabulary}

\author[1]{Dmitry Kobak\,\orcidlink{0000-0002-5639-7209}\,}
\author[1,$\dagger$]{Rita Gonz\'alez-M\'arquez\,\orcidlink{0009-0005-6840-7979}\,}
\author[2,$\dagger$]{Em\H{o}ke-\'Agnes Horv\'at\,\orcidlink{0000-0001-7709-1172}\,}
\author[1,$\dagger$]{Jan Lause\,\orcidlink{0000-0003-0946-412X}\,}

\affil[1]{Hertie Institute for AI in Brain Health, University of T\"ubingen, Germany}
\affil[2]{Northwestern University, Evanston, Illinois, USA}
\affil[$\dagger$]{Alphabetic order}
\affil[ ]{\Letter\: \normalfont{\texttt{dmitry.kobak@uni-tuebingen.de}}}

\usepackage{titling}
\pretitle{\begin{flushleft}\rule{\textwidth}{0.8pt}\end{flushleft} \begin{center}\huge\sc}
\posttitle{\par\end{center} \rule{\textwidth}{0.8pt} \vskip 0.5em}

\usepackage{fancyhdr}
\begin{document}

\twocolumn[
\begin{@twocolumnfalse}
\maketitle

\begin{abstract}
Large language models (LLMs) like ChatGPT can generate and revise text with human-level performance. These models come with clear limitations: they can produce inaccurate information, reinforce existing biases, and be easily misused. Yet, many scientists use them for their scholarly writing. But how wide-spread is such LLM usage in the academic literature? To answer this question for the field of biomedical research, we present an unbiased, large-scale approach: we study vocabulary changes in over 15 million biomedical abstracts from 2010--2024 indexed by PubMed, and show how the appearance of LLMs led to an abrupt increase in the frequency of certain style words. This excess word analysis suggests that at least 13.5\% of 2024 abstracts were processed with LLMs. This lower bound differed across disciplines, countries, and journals, reaching 40\% for some subcorpora. We show that LLMs have had an unprecedented impact on scientific writing in biomedical research, surpassing the effect of major world events such as the Covid pandemic.
\end{abstract}

\vspace{2em}
\end{@twocolumnfalse}
]

\section{Introduction}


When the world changes, human-written text changes. Major events like wars and revolutions affect word frequency distributions in text corpora~\citep{bochkarev2014universals}. The rise and fall of scientific disciplines is traceable in scholarly writing~\citep{hall2008studying,bizzoni2020linguistic}. Do technological advances leave a similar footprint on our writing?

With the release of ChatGPT in November 2022, human writing underwent an unprecedented change: For the first time, a large language model (LLM) was widely available that could generate and revise texts with human-like performance in several domains --- including academia~\citep{ahmed2023future}, where many have hoped that LLMs might lead to more equity~\citep{berdejo2023ai}. Many researchers have since integrated LLMs in their daily writing tasks~\citep{van2023ai} and even co-authored papers with LLMs~\citep{stockel2023chatgpt}. This has led to worries about research integrity, factual mistakes in LLM-generated content~\citep{mittelstadt2023protect,lindsay2023llms,walters2023fabrication,ji2023survey,zhang2023siren}, and misuse of LLMs by so-called paper mills that produce fake publications~\citep{kendall2024risks}. These worries sparked attempts to track the footprint of LLM-assisted writing in scientific texts.

\begin{figure}[t]
    \centering
    \includegraphics[width=\linewidth]{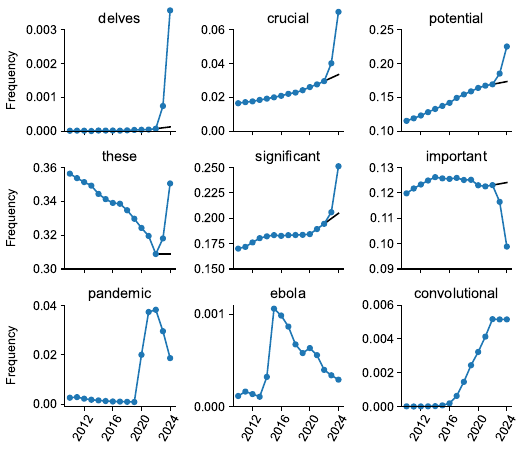}
    \caption{Frequencies of PubMed abstracts containing several example words. Black lines show counterfactual extrapolations from 2021--2022 to 2023--2024. We manually selected words to illustrate the diversity of frequency time courses. The first six words are affected by LLMs; the last three relate to major events that influenced scientific writing and are shown for comparison.}
    \label{fig:examples}
\end{figure}

\begin{figure*}
    \centering
    \includegraphics[width=\textwidth]{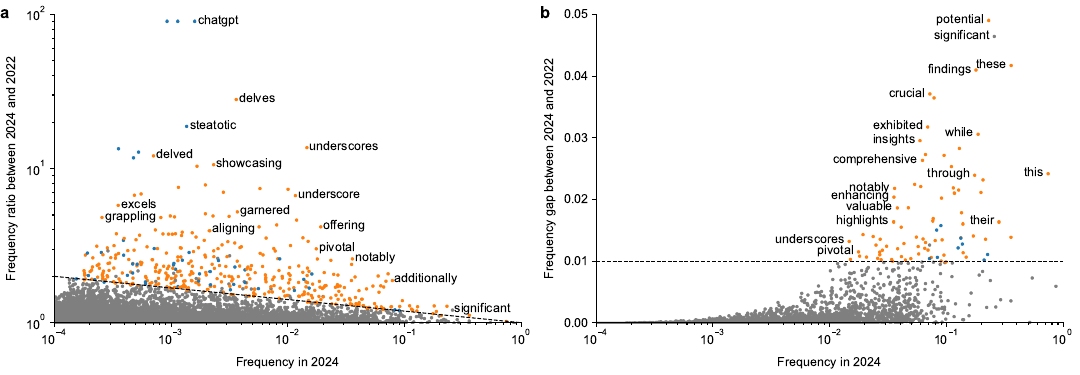}
    \caption{Words showing increased frequency in 2024. \textbf{(a)} Frequencies in 2024 and frequency ratios ($r$). Both axes are on log-scale. Only a subset of points are labeled for visual clarity. The dashed line shows the threshold defining excess words (see text). Words with $r>90$ are shown at $r=90$. Excess words were manually annotated into content words (blue) and style words (orange). \textbf{(b)} The same but with frequency gap ($\delta)$ as the vertical axis. Words with $\delta>0.05$ are shown at $\delta=0.05$.}
    \label{fig:words2024}
\end{figure*}


Recent approaches attempting to quantify the increasing use of LLMs in scientific papers all build on the idea that LLM-written text differs from human-written text~\citep{lazebnik2024detecting, desaire2023distinguishing} and fall in three groups. One group of studies employed  LLM detectors~\citep{tang2024science}, which are blackbox models trained to detect LLM writing based on ground-truth human and LLM texts~\citep{akram2024quantitative,cheng2024have,liu2024towards,picazo2024analysing}. Another group of works explicitly modeled word frequency distribution in scientific corpora as a mixture distribution of texts produced by humans and by LLMs, again estimated using ground-truth human and LLM texts~\citep{liang2024mapping,liang2024monitoring,geng2024chatgpt,astarita2024delving}. The third group of studies relied on lists of marker words, known to be over-used by LLMs, which are typically stylistic words unrelated to the text content~\citep{gray2024chatgpt,liu2024towards,matsui2024delving}.


All of these approaches share one common limitation: they require a ground-truth training set of LLM- and human-written texts. Usually, human-written texts are obtained from pre-LLM years, while LLM-written texts are generated by a set of prompts. This setup can introduce biases~\citep{tang2024science}, as it requires assumptions on which models scientists use for their LLM-assisted writing, and how exactly they prompt them. Furthermore, work based on LLM detector models suffers from their blackbox nature, as it does not allow further interrogation of the results on the word level, which can make their interpretation difficult. Most importantly, all existing work focuses on detecting LLM texts, and none has attempted to systematically compare or relate LLM-induced changes in scientific writing to previous shifts in scholarly texts. This begs the question if the nature and magnitude of the observed changes are comparable to changes that regularly occur due to changing fashions, rising research topics, and global events such as the Covid-19 pandemic~---~or if LLMs impact scientific writing in an unprecedented way. 



Here, we suggest a novel, data-driven approach to track LLM usage without requiring a labeled corpus of human-written versus model-generated texts. We were inspired by studies of excess mortality~\citep{islam2021excess,karlinsky2021tracking,msemburi2023estimates} that looked at the excess of fatalities during the Covid pandemic compared to pre-Covid mortality. We adapt this idea to LLM-induced changes in word usage and track the excess use of words after the release of ChatGPT-like LLMs compared to pre-LLM years. 
Applying this analysis to the corpus of over 15 million 2010--2024 biomedical abstracts from the PubMed library allowed us to track changes in scientific writing over the last decade in the large field of biomedical research. We found that the LLM-induced changes were unprecedented in both quality and quantity.

\section{Results}

\subsection{Excess words indicate widespread LLM usage}

We downloaded all PubMed abstracts until the end of 2024 and used all 15.1~M English-language abstracts from 2010 onwards (after cleaning them from contaminating strings, see Methods). We then computed the matrix of word occurrences that shows which abstracts contain which words, resulting in a 15.1~M $\times$ 273~K sparse binary matrix. For each word and each year, we found the number of abstracts in that year that the word appeared in, and obtained its occurrence frequency $p$ by normalizing with the total number of papers published in that year. Our main analysis focused on 26.7~K words with $p>10^{-4}$ in both 2023 and 2024. With over 1~million abstracts per year, this corresponds to ${>}100$ usages per year. 

Some words strongly increased their occurrence frequency in 2023--2024 (Figure~\ref{fig:examples}). To quantify this increase, we calculated counterfactual expected frequency in 2024 based on the linear extrapolation of word frequencies in 2021 and 2022 (see Methods). Note that we did not use 2023 frequencies for calculating the expected frequency, because they could already have been affected by LLM usage. Comparing the empirical 2024 frequency $p$ with the expected 2024 frequency $q$, we obtained the excess frequency gap $\delta = p-q$ and the excess frequency ratio $r = p/q$ as two measures of excess usage. These two measures are complementary. The frequency gap is well-suited to highlight excess usage of frequent words, while the frequency ratio points to the excess usage of infrequent words. For example, frequency increases from 0.001 to 0.01 and from 0.5 to 0.6 are both noteworthy. Yet, the former frequency increase is captured by a high $r$ value whereas the latter has a high $\delta$ value.

Across all 26.7~K words, we found many with strong excess usage in 2024 (Figure~\ref{fig:words2024}). Less common words with strong excess usage included \textit{delves} ($r=28.0$), \textit{underscores} ($r=13.8$), and \textit{showcasing} ($r=10.7$), together with their grammatical inflections (Figure~\ref{fig:words2024}a). More common words with strong excess usage included \textit{potential} ($\delta=0.052$), \textit{findings} ($\delta=0.041$), and \textit{crucial} ($\delta=0.037$) (Figure~\ref{fig:words2024}b). 

Is this unusual, or do similar frequency changes happen every year? For comparison, we did the same analysis for all years from 2013 to 2023 (Figures~\ref{fig:words2013-14}--\ref{fig:words2021-23}). We found words like \textit{ebola} with $r=9.9$ in 2015 and \textit{zika} with $r=40.4$ in 2017, but from 2013 until 2019, no single word has ever shown excess frequency gap $\delta>0.01$. This changed during the Covid pandemic: in 2020--2022 words like \textit{coronavirus}, \textit{covid}, \textit{lockdown}, and \textit{pandemic} showed very large excess usages (up to $r>1000$ and $\delta=0.06$), in agreement with the observation that the Covid pandemic had an unprecedented effect on biomedical publishing~\citep{gonzalez2024landscape}.

\begin{figure}
    \centering
    \includegraphics[width=\linewidth]{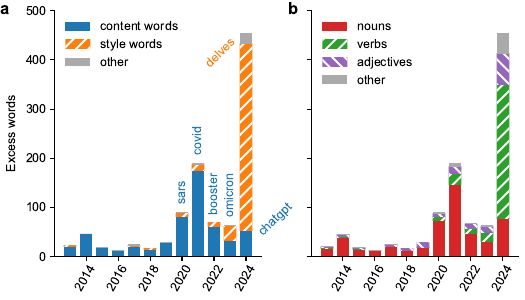}
    \caption{\textbf{(a)} Number of excess words per year, decomposed into the excess content words and excess style words. In each year, we show, as an example, the word with the highest frequency ratio $r$ among excess words with $p>0.0015$ and $r>3$ (in blue). \textbf{(b)} Number of excess words per year, decomposed into nouns, verbs, and adjectives.}
    \label{fig:excess-per-year}
\end{figure}

To compare the size of excess vocabulary between years, we defined as excess words all words with $\delta>0.01$ or 
$\log_{10}r > \frac{\log_{10} 2}{4}\log_{10} p$ where $p$ is frequency in 2024 (see dashed lines in Figure~\ref{fig:words2024}); these thresholds were chosen such that most words in pre-Covid years were well below (Figures~\ref{fig:words2013-14}--\ref{fig:words2021-23}). The number of excess words showed a marked rise during the Covid pandemic (up to 190 words in 2021) followed by an even larger rise (to 454) in 2024 (Figure~\ref{fig:excess-per-year}), roughly one year after ChatGPT was released. Note that this counts grammatical inflections (such as \textit{delve}, \textit{delves}, \textit{delving}, and \textit{delved}; or \textit{mask} and \textit{masks}) multiple times. Counting only unique word lemmas still put 2024 on a clear first place: 343 unique lemmas in 2024 vs. 180 in 2021 (Figure~\ref{fig:excess-per-year-lemmas}).

We manually annotated all 900 unique excess words from 2013--2024 into content words (51.3\%), like \textit{masks} or \textit{convolutional}, and style words (45.2\%), like \textit{intricate} or \textit{notably} (and a small number of ambiguous words, see Methods). The excess vocabulary during the Covid pandemic consisted almost entirely of content words (such as \textit{respiratory}, \textit{remdesivir}, etc.), whereas the excess vocabulary in 2024 consisted almost entirely of style words (Figure~\ref{fig:excess-per-year}a). We also manually assigned part of speech to each excess word. Content words were predominantly nouns (79.2\%), and hence most excess words prior to 2024 were nouns. In contrast, out of all 379 excess style words in 2024, 66\% were verbs and 14\% were adjectives (Figure~\ref{fig:excess-per-year}b). 

\begin{figure}
    \centering
    \includegraphics[width=\linewidth]{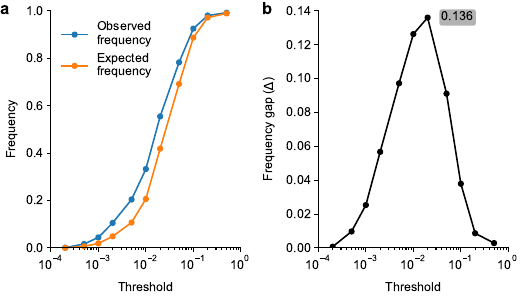}
    \caption{\textbf{(a)} Observed frequency ($P$) and counterfactual expected frequency ($Q$) in 2024 of abstracts containing at least one of the excess style words from 2024 with frequency $p$ below a given threshold. \textbf{(b)} The frequency gap $\Delta=P-Q$ as a function of the threshold.}
    \label{fig:delta-curve}
\end{figure}

\begin{figure*}
    \centering
    \includegraphics[width=\linewidth]{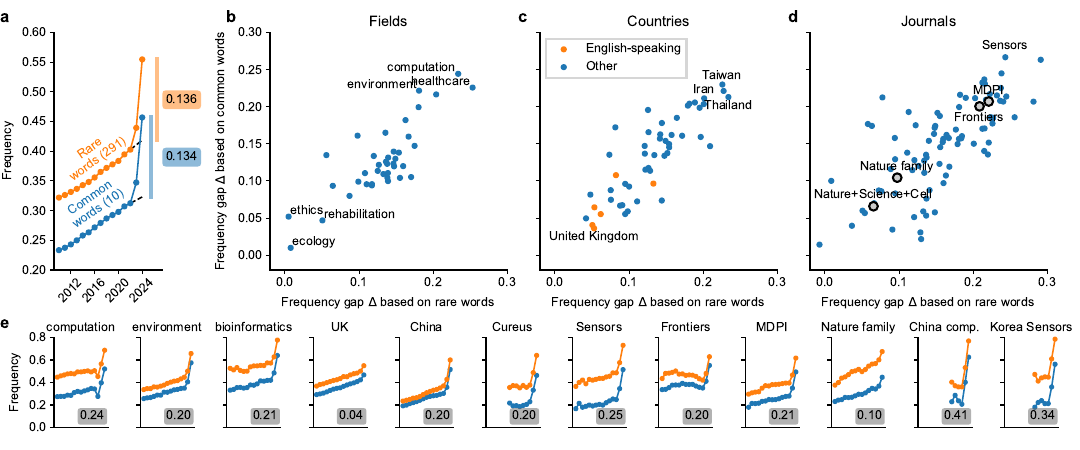}
    \caption{\textbf{(a)} Frequency of abstracts containing at least one word from a given word group. \textbf{(b)} Frequency gap estimates for various fields identified based on journal names \citep{gonzalez2024landscape}. \textbf{(c)} Frequency gap estimates for various countries. \textbf{(d)} Frequency gap estimates for various journals. Gray circles show multiple journals grouped together. \textbf{(e)} Frequencies as in (a) for various PubMed subsets; $\Delta= (\Delta_\mathrm{common} + \Delta_\mathrm{rare})/2$ values are shown.}
    \label{fig:subgroups}
\end{figure*}

\subsection{Combining excess words puts a lower bound on LLM usage}

The unprecedented increase in excess style words in 2024 allows to use them as markers of LLM usage. Each frequency gap $\delta$ for a single marker word gives a lower bound on the fraction of abstracts that went through LLMs in 2024. For example, $\delta=0.052$ for the LLM style marker word \textit{potential} means that in 2024 there were 5.2 percentage points more abstracts containing that word than expected based on the 2021--2022 data, suggesting that at least 5.2\% of all abstracts in 2024 went through an LLM. We reasoned that combining multiple marker words together can increase the lower bound. 

For a given set of marker words $G$, we computed its frequency gap $\Delta_G=P_G - Q_G$  where $P_G$ and $Q_G$ are the observed and the expected fraction of abstracts containing at least one of the words from $G$. As above, we interpret $\Delta_G$ as a lower bound on the LLM usage. Note that this does not assume independence between words in $G$. To avoid a computationally expensive combinatorial search for a set of words with the highest $\Delta_G$, we used two independent heuristics to form two separate sets, focusing on either \textit{rare} or \textit{common} style excess words. 

To create the rare set, we grouped all 2024 excess style words with frequency $p<T$ and computed the frequency gap $\Delta_\mathrm{rare}$ as a function of threshold $T$ (Figure~\ref{fig:delta-curve}). We obtained the highest $\Delta_{\mathrm{rare}}$ value with $T = 0.02$, resulting in a set of 291 words (Figure~\ref{fig:rare-words}). The frequency gap was $\Delta_\mathrm{rare} = 0.136$, putting the lower bound on the LLM usage in 2024 at 13.6\% (Figure~\ref{fig:subgroups}a). Importantly, this is only a lower bound because some of the abstracts that did go through an LLM may fail to contain any of the style words we selected here (see Discussion).

To create the common set, we grouped together excess style words with high individual $\delta$ values, manually adjusting the selection to maximize the frequency gap $\Delta_{\mathrm{common}}$. This led to the following set of 10 words: \textit{across}, \textit{additionally}, \textit{comprehensive}, \textit{crucial}, \textit{enhancing}, \textit{exhibited}, \textit{insights}, \textit{notably}, \textit{particularly}, \textit{within}. This set yielded a very similar frequency gap as the rare words: $\Delta_\mathrm{common} = 0.134$ (Figure~\ref{fig:subgroups}a).

As the rare and the common sets were non-overlapping, both serve as an independent estimate of the lower bound. Averaging the two estimates, we obtained $\Delta = (\Delta_\mathrm{common} + \Delta_\mathrm{rare})/2 = 0.135$ as our final estimate of the lower bound of LLM usage (13.5\%).

For comparison, using the set of four excess content words from 2021, \textit{covid}, \textit{pandemic}, \textit{coronavirus}, and \textit{sars}, yielded frequency gap $\Delta = 0.069$. This shows that the LLM usage in 2024 was at least two times higher than the size of Covid-related literature in 2021.

\subsection{Lower bounds differed between subcorpora}

We performed the same analysis as above by various subgroups of PubMed papers. We computed frequency gaps $\Delta_\mathrm{common}$ and $\Delta_\mathrm{rare}$ for different biomedical fields, affiliation countries, journals, and men and women among the first and the last authors, inferred from their first names (see Methods). Note that we based all our $\Delta$ estimates on the same two sets of excess style words as before. While there may be additional subcorpus-specific excess words, using the same sets of marker words ensures a fair and consistent comparison between the subcorpora.

We found pronounced heterogeneity among most of these categories. Computational fields like computation and bioinformatics showed $\Delta \approx 0.20$ (Figure~\ref{fig:subgroups}b). Among countries, some English-speaking countries like United Kingdom and Australia showed $\Delta \approx 0.05$, while countries like China, South Korea, and Taiwan showed $\Delta \approx 0.20$ (Figure~\ref{fig:subgroups}b). The difference between inferred genders was minor (0.09 for male and 0.07--0.08 for female, both for the first and the last authors).

Among individual journals, we found very high $\Delta$ values, e.g. 0.25 for \textit{Sensors} (an open access journal published by MDPI) and 0.20 for \textit{Cureus} (an open access journal with simplified review process, published by Springer Nature). We analyzed several groups of journals pooled together and found high $\Delta$ for MDPI (0.21) and Frontiers (0.20) journals. For very selective high-prestige journals like \textit{Nature}, \textit{Science}, and \textit{Cell}, and for Nature family journals, $\Delta$ was much lower (0.07 and 0.10 respectively), suggesting that easily-detectable LLM usage was negatively correlated with perceived prestige.

To find subgroups with the strongest effect, we looked at intersections of different groups and found $\Delta=0.34$ for papers from South Korea published in \textit{Sensors} and $\Delta=0.41$ for computation papers from China. An even more fine-grained analysis based on the $t$-SNE embedding \citep{maaten2008visualizing} of 2022--2024 papers showed some areas with local $\Delta\approx 0.50$ (Figure~\ref{fig:tsne}), such as, for example, a cluster of papers on deep-learning-based object detection, with predominantly Chinese affiliations and with the majority published in MDPI's \textit{Sensors}.

\section{Discussion}

\paragraph{Summary}

In this paper, we leveraged excess word usage to show how LLMs have affected scientific writing in biomedical research. We found that the effect was unprecedented in quality and quantity: hundreds of words have abruptly increased their frequency after ChatGPT-like LLMs became available. In contrast to previous shifts in word popularity, the 2023--2024 excess words were not content-related nouns, but rather style-affecting verbs and adjectives that LLMs prefer. 

Our analysis is performed on the corpus level and cannot identify individual abstracts that may have been processed by an LLM. Still, the following examples from three real 2023 abstracts illustrate the LLM-style flowery language: 
\begin{itemize}
    \item \textit{By \textbf{meticulously} \textbf{delving} into the \textbf{intricate} web connecting [...] and [...], this \textbf{comprehensive} chapter takes a deep dive into their involvement as \textbf{significant} risk factors for [...].} 

    \item \textit{A \textbf{comprehensive} grasp of the \textbf{intricate} \textbf{interplay} between [...] and [...] is \textbf{pivotal} for effective therapeutic strategies.} 

    \item \textit{Initially, we \textbf{delve} into the \textbf{intricacies} of [...], \textbf{accentuating} its indispensability in cellular physiology, the enzymatic labyrinth governing its flux, and the \textbf{pivotal} [...] mechanisms.}
\end{itemize}
Our analysis of the excess frequency of such LLM-preferred style words suggests that at least 13.5\% of 2024 PubMed abstracts were processed with LLMs. With ${\sim}1.5$~million papers being currently indexed in PubMed per year, this means that LLMs assist in writing at least 200\,000 papers per year. This estimate is based on LLM marker words that showed large excess usage in 2024, which strongly suggests these words are preferred by LLMs like ChatGPT that became popular by that time. Importantly, this is only a lower bound: abstracts not using any of the LLM marker words are not contributing to our estimates, so the true fraction of LLM-processed abstracts is likely higher.

\paragraph{Interpretation and limitations}


Our estimated lower bound on LLM usage ranged from below 5\% to over 40\% across different PubMed-indexed research fields, affiliation countries, and journals. This heterogeneity could correspond to actual differences in LLM adoption. For example, the high lower bound on LLM usage in computational fields (20\%) could be due to computer science researchers being more familiar with and willing to adopt LLM technology. In non-English speaking countries, LLMs can help authors with editing English texts, which could justify their extensive use. Finally, authors publishing in journals with expedited and/or simplified review processes might be grabbing for LLMs to write low-effort articles. 

However, the heterogeneity in lower bounds could also point to other factors beyond actual differences in LLM adoption. First, it could highlight non-trivial discrepancies in how authors of different linguistic backgrounds censor suggestions from writing assistants, thereby making the use of LLMs non-detectable for word-based approaches like the one we developed here. It is possible that native and non-native English speakers actually use LLMs equally often, but native speakers may be better at noticing and actively removing unnatural style words from LLM outputs. Our method would not be able to pick up the increased frequency of such more advanced LLM usage. Second, publication timelines in computational fields are often shorter than in many biomedical or clinical areas, meaning that any potential increase in LLM usage can be detected earlier in computational journals. Third, the same is true for journals and publishers with faster turnaround times than thoroughly reviewed, high-prestige journals. Our method can easily be used to reevaluate these results after a couple of publication cycles in all fields and journals. We expect the lower bounds documented here to increase with these longer observation windows. 

Given these potential explanations for the heterogeneity in the lower bound of LLM use for scientific editing, our results indicate widespread usage in most PubMed-indexed fields, countries, and journals, including the most prestigious ones. We argue that the true LLM usage in biomedical publishing may be closer to the highest lower bounds we observed, as those may be corpora where LLM usage is the most na\"ive and the easiest to detect. These estimates are above 30\%, which is in line with recent surveys on researchers' use of LLMs for manuscript writing~\citep{van2023ai}. Our results show how those self-reported behaviors translate into real-world LLM usage in final publications.



Finally, while our approach can detect unexpected lexical changes, it cannot separate different causes behind those changes, like multiple emerging topics or multiple emerging writing style changes. For example, our approach cannot distinguish word frequency increase due to direct LLM usage from word frequency increase due to people adopting LLM-preferred words and borrowing them for their own writing. For spoken language, there is emerging evidence for such influence of LLMs on human language usage~\citep{yakura2024empirical}. However, we hypothesize that this effect is much smaller and much slower. Similarly, we cannot distinguish the influence of different LLMs. 

\paragraph{Related work}

Our results go beyond other studies on detecting LLM fingerprints in academic writing. \citet{gray2024chatgpt} described a $2$-fold increase in frequency for the words \textit{intricate} and \textit{meticulously} in 2023, while \citet{liang2024mapping} identified \textit{pivotal}, \textit{intricate}, \textit{showcasing}, and \textit{realm} as the top LLM-preferred words based on a corpus of LLM-generated text. Our study is the first to perform a systematic search for LLM marker words based on excess usage in published scientific texts. We found 379 style words with highly elevated frequencies in 2024, 
and indeed all the above examples appear in our list. 

Some studies have reported differences in estimated LLM usage between English- and non-English-speaking countries~\citep{cheng2024have,liu2024towards,picazo2024analysing}, academic fields~\citep{akram2024quantitative}, and publishing venues. For example, \citet{liang2024mapping} estimated the fraction of LLM-assisted papers in early 2024 to vary between 7\% for Nature Portfolio papers and 17\% for computer science preprints. Importantly, our analysis is based on 5--200 times more papers per year than these prior works, which allowed us to study LLM adoption with greater statistical power and across a much larger diversity of countries, fields, and journals.

Importantly, our approach has two conceptual advantages over previous work: First, all prior studies relied on ground-truth LLM-generated and human-written scientific texts. Such datasets can easily be biased~\citep{tang2024science}, with no guarantee that the corpus of LLM-generated texts is representative of all LLM use cases occurring in actual scholarly practice. In contrast, our analysis avoids this limitation by detecting emerging LLM fingerprints directly from published abstracts. Second, our approach is not restricted to LLM usage and can be applied to abstracts from previous years. This allowed us to put the LLM-induced changes in scientific writing into a historic context (e.g., comparing the influence of LLMs to the influence of the Covid-19 pandemic), and to conclude that these changes are without precedent.

\paragraph{Implications and policies}

What are the implications of this ongoing revolution in scientific writing? Scientists use LLM-assisted writing because LLMs can improve grammar, rhetoric, and overall readability of their texts, help translate to English, and quickly generate summaries~\citep{van2024adapted,zhang2024benchmarking}. However, LLMs are infamous for making up references~\citep{walters2023fabrication}, providing inaccurate summaries~\citep{tang2024tofueval,kim2024fables}, and making false claims that sound authoritative and convincing~\citep{mittelstadt2023protect,ji2023survey,zhang2023siren,zheng2023chatgpt}. While researchers may notice and correct factual mistakes in LLM-assisted summaries of their own work, it may be harder to spot errors in LLM-generated literature reviews or discussion sections. 

Furthermore, LLMs can mimic biases and other deficiencies from their training data~\citep{bender2021dangers,navigli2023biases,bai2024measuring,choudhury2023generative}, or even outright plagiarise~\citep{mccoy2023much}. This makes LLM outputs less diverse and novel than human-written text~\citep{padmakumar2023does,alvero2024large}. Such homogenisation can degrade the quality of scientific writing. For instance, all LLM-generated introductions on a certain topic might sound the same and would contain the same set of ideas and references, thereby missing out on innovations~\citep{nakadai2023ai} and exacerbating citation injustice~\citep{dworkin2020citing}.
Even worse, it is likely that malign actors such as paper mills will employ LLMs to produce fake publications~\citep{kendall2024risks}.

Our work shows that LLM usage for scientific writing is on the rise despite these substantial limitations. How should the academic community deal with this development? Some have suggested to use retrieval-augmented LLMs that provide verifiable facts from trusted sources~\citep{lewis2020retrieval,borgeaud2022improving,ahmed2023future} or let the user provide all relevant facts to the LLM in order to protect scientific literature from accumulating subtle inaccuracies~\citep{mittelstadt2023protect}. Others think that for certain tasks like peer reviewing, LLMs are ill-suited and should not be used at all~\citep{lindsay2023llms}. As a result, publishers and funding agencies have put out various policies, banning LLMs in peer review~\citep{kaiser2023funding,brainard2023scientists}, as co-authors~\citep{thorp2023chatgpt}, or undisclosed resource of any kind~\citep{brainard2023scientists}. Data-driven and unbiased analyses like ours can be helpful to monitor whether such policies are ignored or adhered to in practice.

In conclusion, our work showed that the effect of LLM usage on scientific writing is truly unprecedented and outshines even the drastic changes in vocabulary induced by the Covid-19 pandemic. This effect will likely become even more pronounced in the future, as one can analyze more publication cycles and LLMs are likely to increase in adoption. At the same time, LLM usage can be well-disguised and hard to detect, so the true extent of their adoption is likely already higher than what we measured. This trend calls for a reassessment of current policies and regulations around the use of LLMs for science.  Our analysis can inform the necessary debate around LLM policies by providing a measurement method for LLM usage that is urgently needed~\citep{brinkmann2023machine,heersmink2024use}. Our excess word approach could help to track future LLM usage, including scientific (grant applications and peer review) and non-scientific (news articles, social media, prose) use cases.
We hope that future work will meticulously delve into tracking LLM usage more accurately and assess which policy changes are crucial to tackle the intricate challenges posed by the rise of LLMs in scientific publishing. 

\section{Materials and Methods}

\subsection{Dataset}

We used the annual PubMed snapshot released in the beginning of 2025 (\url{https://ftp.ncbi.nlm.nih.gov/pubmed/baseline/}), containing files from \texttt{pubmed25n0001.xml.gz} to \texttt{pubmed25n1274.xml.gz}. We parsed the XML files as described in our prior work \citep{gonzalez2024landscape}, applying the same filtering criteria, to extract the abstract texts and some metadata, keeping only complete English-language abstracts with length 250--4000 characters. This resulted in 24\,814\,136 abstracts.
We then only analyzed papers with publication years from 2010 to 2024, giving us 15\,103\,888 abstracts for analysis.

For each paper, we defined the affiliation country as the country of the first affiliation of the first author. We assigned papers to the 39 fields taken from \citet{gonzalez2024landscape} based on the journal names: for example, all papers from all journals containing the word \textit{neuroscience} (such as \textit{The Journal of Neuroscience} or \textit{Nature Neuroscience}) were assigned to the `neuroscience' field. We used the first names of the first and the last authors to infer their genders via the \texttt{gender} package \citep{genderpackage}. Our gender inference aims to capture perceived gender based on first name and is only approximate. The inference model has clear shortcomings, including limited US-based training data. Moreover, some first names are inherently gender-ambiguous. We were able to infer genders only for 55\% of the authors. See \citet{gonzalez2024landscape} for further details.

\subsection{Pre-processing}

Many abstracts in PubMed data contain strings, usually either in the beginning or in the end, that are not technically part of the abstract text. This can be, for example, ``Communicated by:'' followed by the name of the editor; or ``Copyright \copyright'' followed by the name of the publisher; or ``How to cite this article:'' followed by the citation string. Such strings often appear in abstracts from a particular journal starting from a particular year, and in this case are picked up by our analysis of excess words.

We spent substantial effort to clean the abstracts from all such contaminating strings, using over 100 regular expressions to find and eliminate them. Overall, 286\,744 abstracts were affected by our cleaning procedure. We have also entirely erased 3\,514 abstracts of errata, corrigenda, correction, or retraction notices (identified based on titles).

We then computed a binary word occurrence matrix using \texttt{CountVectorizer(binary=True, min\_df=1e-6)} from Scikit-learn~\citep{pedregosa2011scikit}, obtaining a $15\,103\,888 \times 362\,441$ sparse matrix. We focused the subsequent analysis on $273\,112$ words consisting of at least four letters and composed only out of the 26 letters of English alphabet. Note that different strings (e.g. \textit{mask} and \textit{masks}) were treated as two distinct words.

\subsection{Statistical analysis}

To avoid possible divisions by zero, all frequencies were always computed as $p = (a + 1) / (b + 1)$, where $a$ is the number of abstracts in a given year containing a given word, and $b$ is the total number of abstracts in that year.

When computing excess words in year $Y$, we only looked at words with frequencies above $10^{-4}$ both in year $Y$ and $Y-1$. To do the linear extrapolation, we took the frequencies $p_{-3}$ in year $Y-3$ and $p_{-2}$ in year $Y-2$ and computed the counterfactual projection $q = p_{-2} + 2\cdot\max\{p_{-2} - p_{-3}, 0\}$. This way, $q$ was always at least as large as $p_{-2}$ (see Figure~\ref{fig:examples}), resulting in conservative estimates of $r = p/q$ and $\delta = p - q$.

\subsection{Word annotations and lemmatisation}

We identified 900 unique excess words (surpassing our thresholds on $r$ or $\delta$) from 2013 to 2024. Some of these words showed excess usage in multiple years. We sorted the list alphabetically and annotated them as content and style words while being blinded to the year in which they were selected as excess words. We assigned parts of speech (nouns, adjectives, verbs, etc.) in the same way. In case of doubt, the words were discussed between the authors. When we were not certain whether a word was content or style because of ambiguous usage, we did not label this word as either content or style.

To lemmatise the words for Figure~\ref{fig:excess-per-year-lemmas}, we used \texttt{WordNetLemmatizer()} from the Python NLTK library~\citep{bird2009natural}, and manually added several lemmas such as \textit{chatbots}$\to$\textit{chatbot} and \textit{circrnas}$\to$\textit{circrna}. We also converted British spelling to the US spelling using a dictionary from \url{https://github.com/hyperreality/American-British-English-Translator}.

\subsection{Subgroup analysis}

For the analysis presented in Figure~\ref{fig:subgroups} we separately analysed the following subgroups: 50 affiliation countries with the most papers in our dataset; 100 journals with the most papers in our dataset in 2024; all 39 fields taken from \citet{gonzalez2024landscape}; male and female inferred genders of the first and of the last authors. 

We also analyzed several groups of journals pooled together: (1) \textit{Nature}, \textit{Science}, and \textit{Cell}; (2) 31 specialized Nature family journals established in 2018 or earlier (from \textit{Nature Aging} to \textit{Nature Sustainability}); (3) all Frontiers journals called \texttt{Frontiers *}; (4) all journals published by MDPI, selected based on their PubMed names \texttt{* (Basel, Switzerland)}.

Subgroups were assigned $\Delta$ values only if they contained at least 300 papers in each year from 2018 to 2023.

\subsection{LLM usage}

We did not use any LLMs for writing or editing the manuscript.

\subsection{Data and code availability}

Our analysis code in Python is available at \url{https://github.com/berenslab/llm-excess-vocab}. The $362\,442 \times 15$ matrix of yearly word occurrences (for each word and year, the number of abstracts in that year containing that word; the additional last row contains the total number of abstracts in that year) is available in our Github repository as a \texttt{csv.gz} file. Our content/style annotations are also available there as a \texttt{csv} file. The original PubMed data are openly available for bulk download here: \url{https://pubmed.ncbi.nlm.nih.gov/download}.

\section*{Author contributions}

Conceived and designed the experiments: all authors. Performed the experiments: DK. Analyzed the data: DK. Contributed materials/analysis tools: DK, RGM. Wrote the paper: all authors.

\section*{Acknowledgements}

This work was conceived at the Dagstuhl seminar 24122 supported by the Leibniz Center for Informatics. This work was funded by the Deutsche Forschungsgemeinschaft (KO6282/2-1) and the Gemeinn\"{u}tzige Hertie-Stiftung. The authors thank the International Max Planck Research School for Intelligent Systems (IMPRS-IS) for supporting RGM. DK is a member of the Germany’s Excellence cluster 2064 ``Machine Learning --- New Perspectives for Science'' (EXC 390727645). 
E\'AH's work is supported by NSF CAREER Grant No IIS-1943506. We thank \textit{The Economist} coverage for inspiring us to use stacked bars in Figure~\ref{fig:excess-per-year}.

\bibliographystyle{plainnat}
\bibliography{references.bib}

\clearpage

\onecolumn

\section*{Supplementary Figures}
\renewcommand{\thefigure}{S\arabic{figure}}
\setcounter{figure}{0}  
\renewcommand{\thetable}{S\arabic{table}}
\setcounter{table}{0} 

\begin{figure}[h]
    \centering
    \includegraphics[width=\textwidth]{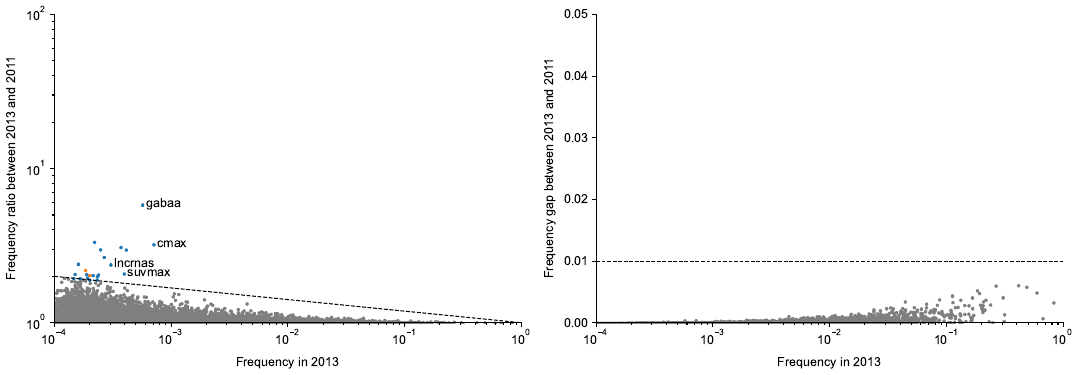}
    \includegraphics[width=\textwidth]{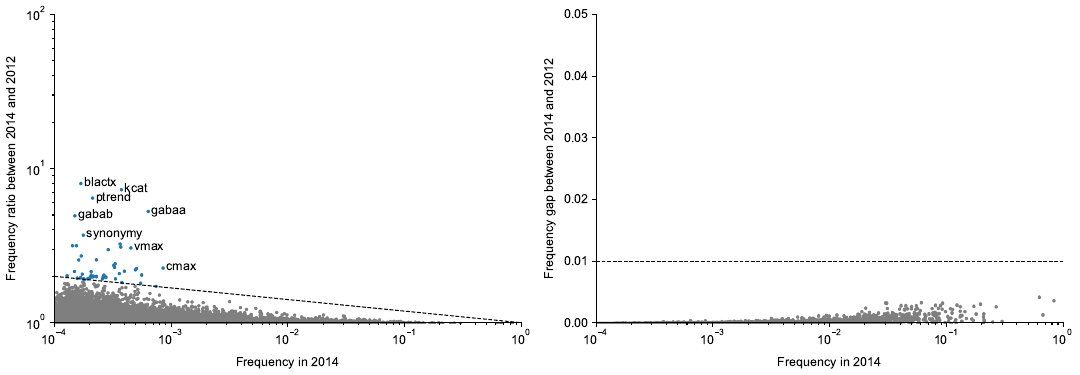}
    \caption{Excess words in 2013 and 2014. See Figure~\ref{fig:words2024} for explanations.}
    \label{fig:words2013-14}
\end{figure}

\begin{figure}
    \centering
    \includegraphics[width=\textwidth]{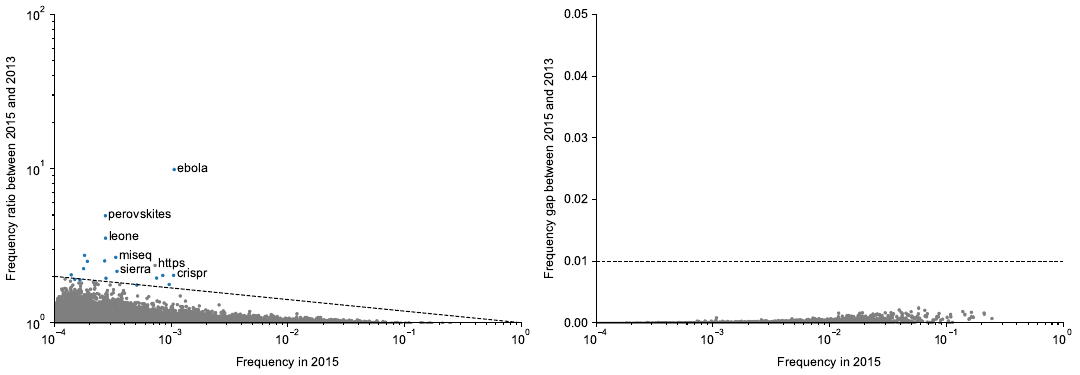}
    \includegraphics[width=\textwidth]{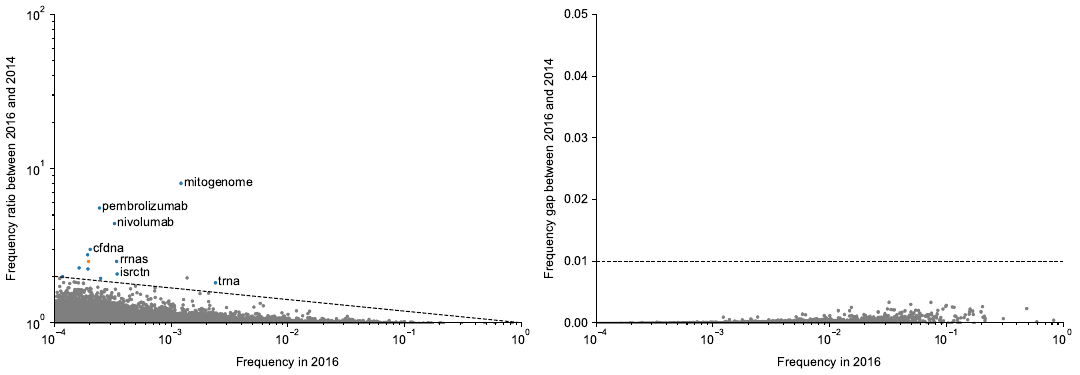}
    \includegraphics[width=\textwidth]{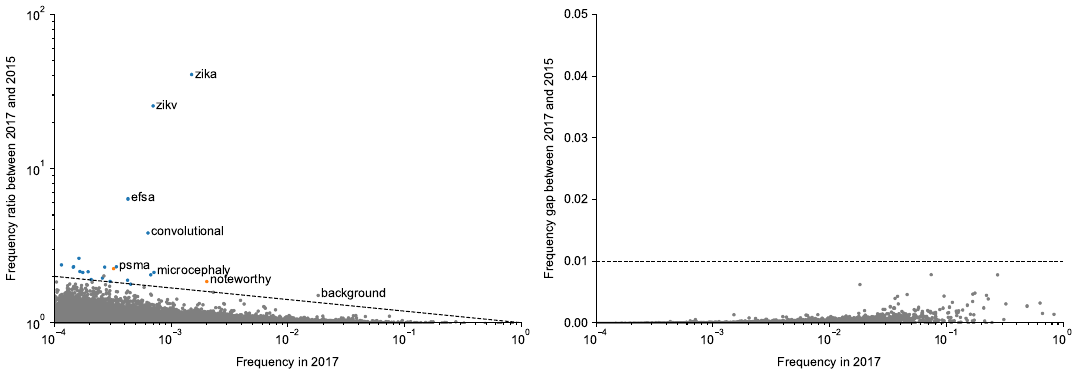}
    \caption{Excess words in 2015--2017. See Figure~\ref{fig:words2024} for explanations. Note that the excess word \textit{mitogenome} (2016) is not shown in Figure~\ref{fig:excess-per-year}a because it was mostly used in a single journal (\textit{Mitochondrial DNA}) and so we considered it not representative of the entire biomedical literature.}
    \label{fig:words2015-17}
\end{figure}

\begin{figure}
    \centering
    \includegraphics[width=\textwidth]{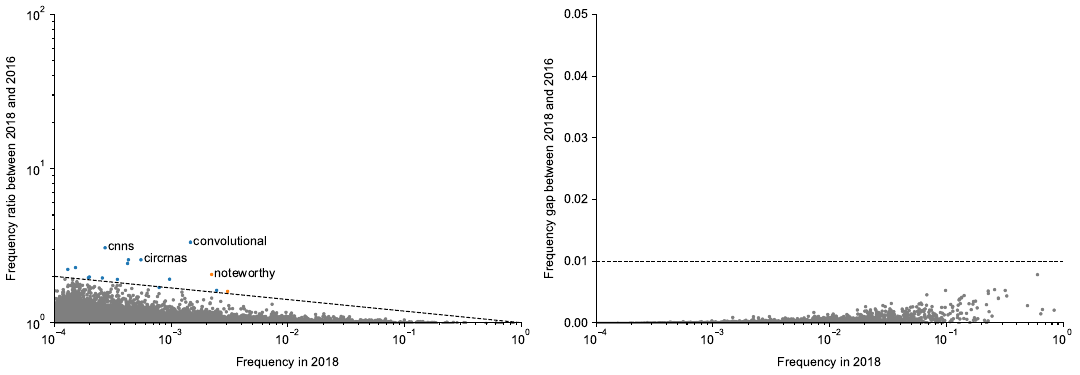}
    \includegraphics[width=\textwidth]{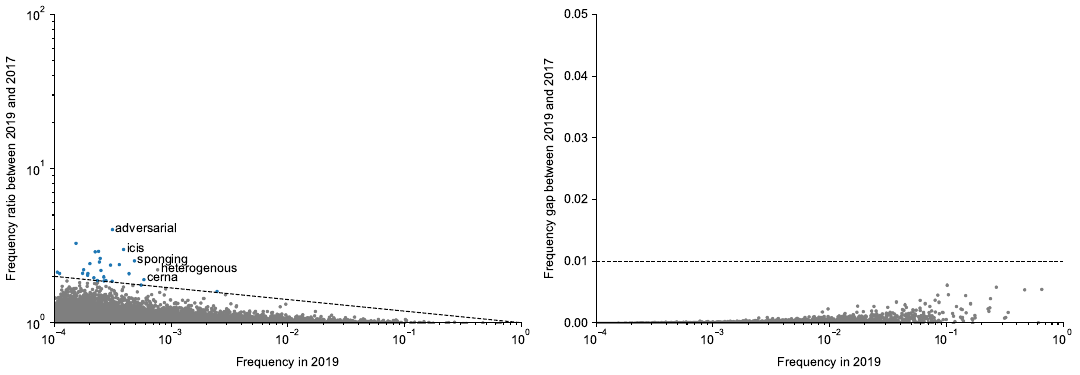}
    \includegraphics[width=\textwidth]{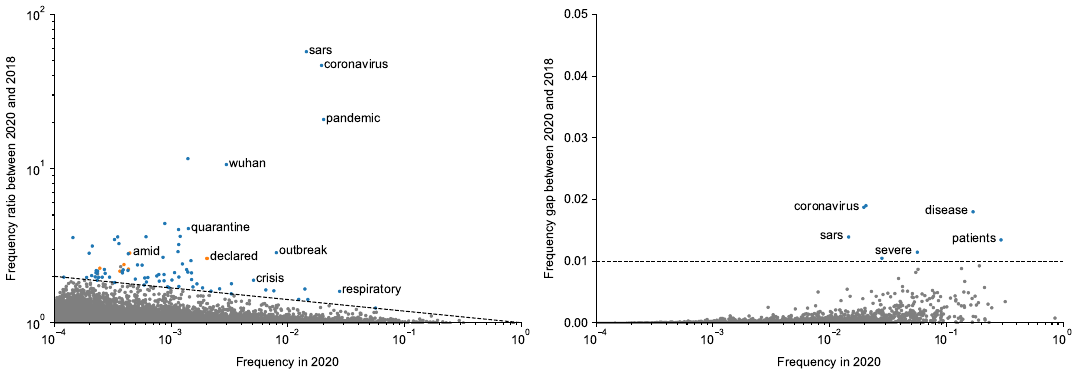}
    \caption{Excess words in 2018--2020. See Figure~\ref{fig:words2024} for explanations.}
    \label{fig:words2018-20}
\end{figure}

\begin{figure}
    \centering
    \includegraphics[width=\textwidth]{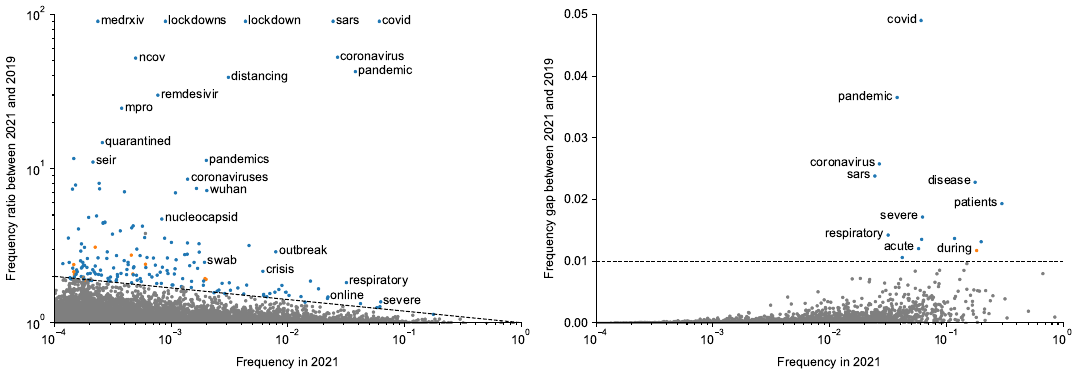}
    \includegraphics[width=\textwidth]{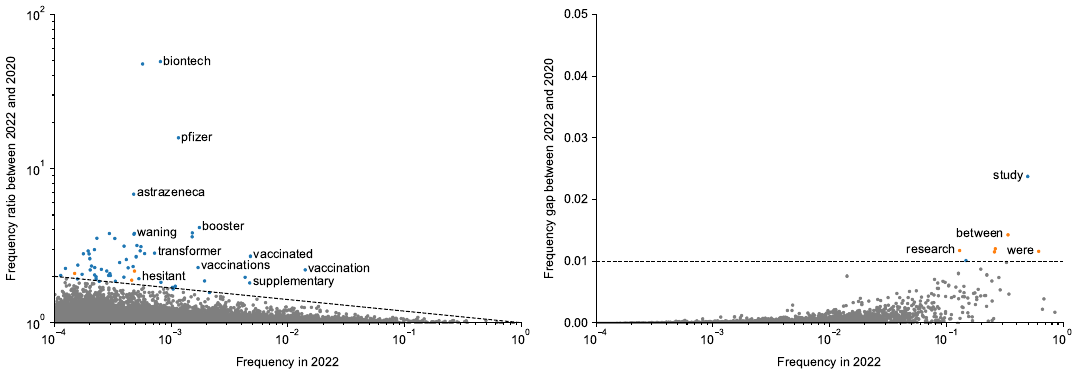}
    \includegraphics[width=\textwidth]{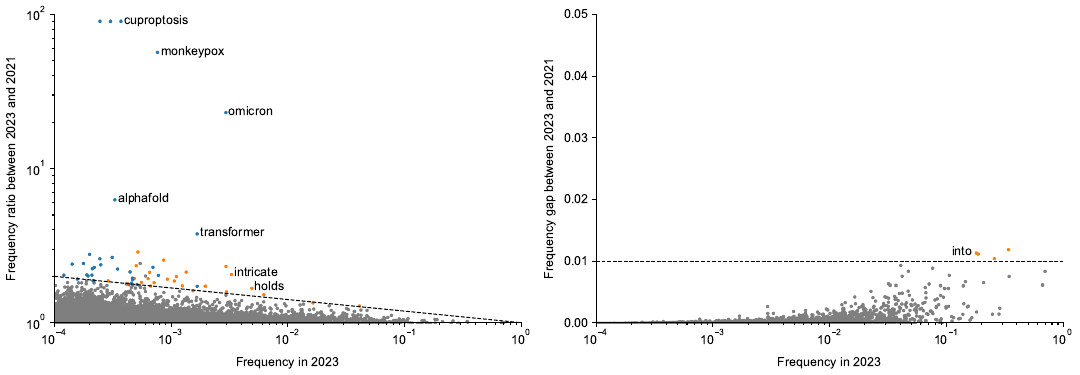}
    \caption{Excess words in 2021--2023. See Figure~\ref{fig:words2024} for explanations.}
    \label{fig:words2021-23}
\end{figure}

\begin{figure}
    \centering
    \includegraphics[width=.48\linewidth]{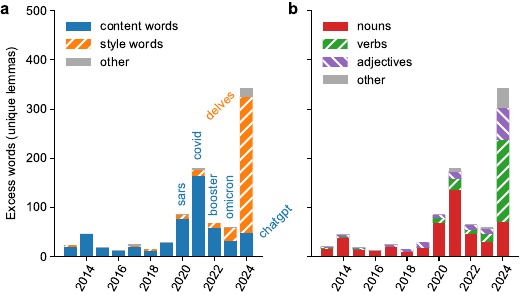}
    \caption{The same as Figure~\ref{fig:excess-per-year} but showing the number of unique lemmas of excess words, instead of the number of excess words. This counts \textit{delve}, \textit{delves}, \textit{delving}, and \textit{delved} (or \textit{mask} and \textit{masks}) only once.}
    \label{fig:excess-per-year-lemmas}
\end{figure}

\begin{figure}
    \texttt{accentuates, acknowledges, acknowledging, addresses, adept, adhered, adhering, advancement, advancements, advancing, advocates, advocating, affirming, afflicted, aiding, akin, align, aligning, aligns, alongside, amidst, assessments, attains, attributed, augmenting, avenue, avenues, bolster, bolstered, bolstering, broader, burgeoning, capabilities, capitalizing, categorized, categorizes, categorizing, combating, commendable, compelling, complicates, complicating, comprehending, comprising, consequently, consolidates, contributing, conversely, correlating, crafted, crafting, culminating, customizing, delineates, delve, delved, delves, delving, demonstrating, dependability, dependable, detailing, detrimentally, diminishes, diminishing, discern, discerned, discernible, discerning, displaying, disrupts, distinctions, distinctive, elevate, elevates, elevating, elucidate, elucidates, elucidating, embracing, emerges, emphasises, emphasising, emphasize, emphasizes, emphasizing, employing, employs, empowers, emulating, emulation, enabling, encapsulates, encompass, encompassed, encompasses, encompassing, endeavors, endeavours, enduring, enhancements, enhances, ensuring, equipping, escalating, evaluates, evolving, exacerbating, examines, exceeding, excels, exceptional, exceptionally, exerting, exhibiting, exhibits, expedite, expediting, exploration, explores, facilitated, facilitates, facilitating, featuring, formidable, fostering, fosters, foundational, furnish, garnered, garnering, gauged, grappling, groundbreaking, groundwork, harness, harnesses, harnessing, heighten, heightened, hinder, hinges, hinting, hold, holds, illuminates, illuminating, imbalances, impacting, impede, impeding, imperative, impressive, inadequately, incorporates, incorporating, influencing, inherent, initially, innovative, inquiries, integrates, integrating, integration, interconnectedness, interplay, intricacies, intricate, intricately, introduces, invaluable, investigates, involves, juxtaposed, leverages, leveraging, maintaining, merges, methodologies, meticulous, meticulously, multifaceted, necessitate, necessitates, necessitating, necessity, notable, noteworthy, nuanced, nuances, offering, optimizing, orchestrating, outlines, overlook, overlooking, paving, persist, pinpoint, pinpointed, pinpointing, pioneering, pioneers, pivotal, poised, pose, posed, poses, posing, predominantly, preserving, pressing, promise, pronounced, propelling, realm, realms, recognizing, refine, refines, refining, remarkable, renowned, revealing, reveals, revolutionize, revolutionizing, revolves, scrutinize, scrutinized, scrutinizing, seamless, seamlessly, seeks, serves, serving, shaping, shedding, showcased, showcases, showcasing, signifying, solidify, spanned, spanning, spurred, stands, stemming, strategically, streamline, streamlined, streamlines, streamlining, struggle, substantiated, substantiates, surged, surmount, surpass, surpassed, surpasses, surpassing, swift, swiftly, thorough, transformative, typically, ultimately, uncharted, uncovering, underexplored, underscore, underscored, underscores, underscoring, unexplored, unlocking, unparalleled, unraveling, unveil, unveiled, unveiling, unveils, uphold, upholding, urging, utilizes, varying, versatility, warranting, yielding}
    \caption{All 291 excess style words in 2024 with frequency below 0.02, see Figure~\ref{fig:delta-curve}.}
    \label{fig:rare-words}
\end{figure}

\begin{figure}
    \centering
    \includegraphics[width=\linewidth]{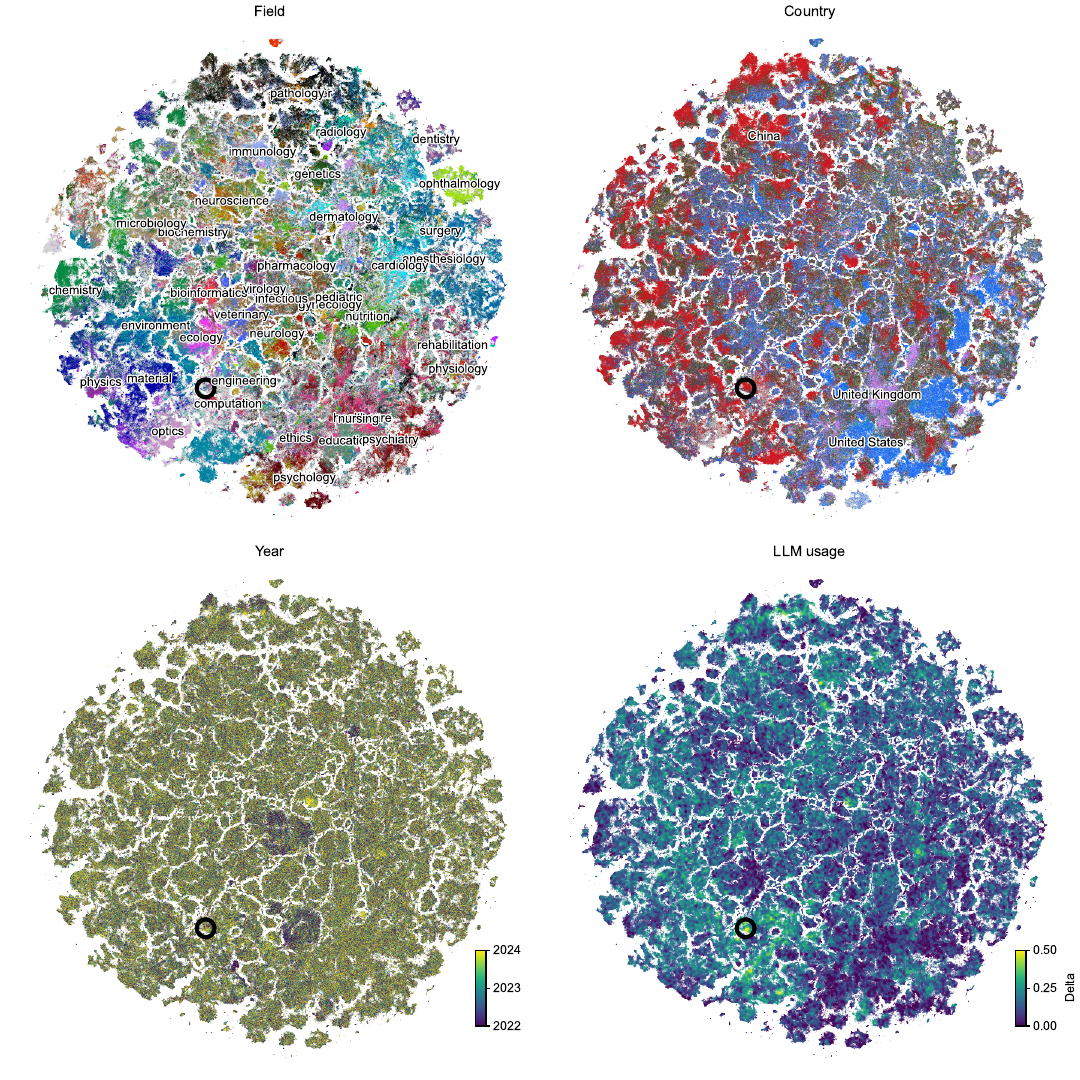}
    \caption{A 2D visualisation of all 4\,109\,080 abstracts from 2022 to 2024. Following \citet{gonzalez2024landscape}, we used the \texttt{[SEP]} token of the pre-trained PubMedBERT model \citep{gu2021domain} to obtain 768-dimensional vector representation of each abstract, and then used $t$-SNE \citep{maaten2008visualizing} to visualize the results in 2D. Top left: colored by field. Labels are positioned at the peak density of each field. Top right: colored by the affiliation country. Bottom left: colored by publication year. Bottom right: colored by local LLM usage. The local LLM usage was defined for each paper based on its nearest neighbors in 2D. Within this local neighborhood, we computed $\Delta_\text{rare}$ and $\Delta_\text{common}$ (see main text for details): For the 100 neighboring 2022 abstracts (and for the 100 neighboring 2024 abstracts), we found the fraction of abstracts containing at least one of the rare  (and at least one of the common) excess words. Taking the difference between years yielded $\Delta_\text{rare}$ and $\Delta_\text{common}$, which we averaged to obtain the final $\Delta$ value. This was done separately for each of the embedded abstracts. Note that the regions with high $\Delta$ do not consist of only 2024 papers (third panel), but rather of a mix of 2022--2024 papers. This means that our embedding model does not group abstracts based on LLM-associated keywords, but rather by topic. Black circle in all panels highlights one example area with $\Delta\approx 0.50$; it consists of $\sim$400 papers on deep-learning-based object detection, with predominantly Chinese affiliations (83.4\%) and with the majority published in \textit{Sensors} (56.3\%). Note: a cluster of papers from 2022 (dark blue in the lower-left panel) to the lower-right of the circle contains papers with predominantly Chinese affiliations (79.9\%), with many (48.7\%) coming from the four journals from the \textit{Hindawi} publisher that were all shut down in 2023 as ``heavily compromised by paper mills'' \citep{retractionwatch2023hindawi}; 28.8\% of the papers in this island have been retracted.}
    \label{fig:tsne}
\end{figure}

\end{document}